\documentclass[conference, 9pt]{IEEEtran}
\IEEEoverridecommandlockouts
\usepackage{cite}
\usepackage{amsmath,amssymb,amsfonts}
\usepackage{algorithmic}
\usepackage{graphicx}
\usepackage{textcomp}
\usepackage{subfig}
\usepackage{subcaption}
\usepackage{xcolor}
\usepackage{url}
\usepackage{array}
\usepackage{booktabs}
\usepackage[normalem]{ulem}
\def\BibTeX{{\rm B\kern-.05em{\sc i\kern-.025em b}\kern-.08em
    T\kern-.1667em\lower.7ex\hbox{E}\kern-.125emX}}

\newcommand{\chang}[1]{\textcolor{black}{#1}}

\newcolumntype{M}[1]{>{\centering\arraybackslash}m{#1}}

\usepackage{fancyhdr}
\usepackage{ragged2e}  

\fancypagestyle{firstpage}{
    \fancyhf{} 
    \setlength{\footskip}{28pt}  
    \fancyfoot[C]{\justifying \footnotesize{\copyright 2025 IEEE. Personal use of this material is permitted. However, permission to reprint/republish this material for advertising or promotional purposes or for creating new collective works for resale or redistribution to servers or lists, or to reuse any copyrighted component of this work in other works, must be obtained from the IEEE.}} 
}

\begin{document}

\title{Investigating the Sensitivity of Pre-trained Audio Embeddings to Common Effects
\thanks{$^*$Equal contribution.
This work was partly funded by the European Union (ERC, HI-Audio, 101052978). Views and opinions expressed are however those of the author(s) only and do not necessarily reflect those of the European Union or the European Research Council. Neither the European Union nor the granting authority can be held responsible for them.
}
}

\author{
\IEEEauthorblockN{ \qquad Victor Deng$^{1,2 \ *}$ \qquad Changhong Wang$^{1 \ *}$ \qquad Gaël Richard$^{1}$ \qquad Brian McFee$^{3}$}

\IEEEauthorblockA{
\textit{$^{1}$LTCI, Télécom Paris, Institut Polytechnique de Paris, France} \\
\textit{$^{2}$Département d'Informatique, École Normale Supérieure, Paris, France}\\
\textit{$^{3}$Music and Audio Research Laboratory, New York University, USA}\\
}
}

\maketitle
\thispagestyle{firstpage}

\begin{abstract}
In recent years, foundation models have significantly advanced data-driven systems across various domains. Yet, their underlying properties, especially when functioning as feature extractors, remain under-explored.
In this paper, we investigate the sensitivity to audio effects of audio embeddings extracted from widely-used foundation models, including OpenL3, PANNs, and CLAP. We focus on audio effects as the source of sensitivity due to their prevalent presence in large audio datasets.
By applying parameterized audio effects (gain, low-pass filtering, reverberation, and bitcrushing), we analyze the correlation between the deformation trajectories and the effect strength in the embedding space. 
We propose to quantify the dimensionality and linearizability of the deformation trajectories induced by audio effects using canonical correlation analysis. We find that there exists a direction along which the
embeddings move monotonically as the audio effect strength increases, but that the subspace containing the displacements is generally high-dimensional.
This shows that pre-trained audio embeddings do not globally linearize the effects. Our empirical results on instrument classification downstream tasks confirm that projecting out the estimated deformation directions cannot generally improve the robustness of pre-trained embeddings to audio effects.
\end{abstract}


\section{Introduction}

The development of foundation models has marked a shift towards large-scale, general-purpose artificial intelligence. 
These models are often trained on vast amounts of data, making them particularly valuable as feature extractors in transfer learning settings.
One popular and effective approach is to leverage features extracted from these models, also called pre-trained embeddings, for downstream tasks with limited data.
Despite their widespread use, there is a lack of research advancing our understanding of these foundation models.  
Many questions remain unanswered, such as what the embeddings represent, what their invariance properties are, and which embedding we should use for a given task.
In this paper, we investigate the sensitivity of pre-trained audio embeddings to common audio effects.

A few prior studies have pointed in this direction, but with a limited scope. 
The most related work is~\cite{srivastava2022study}, where the authors explored the sensitivity of two pre-trained audio embeddings (OpenL3 and YAMNet) to microphone channel effects.
They introduced three distance metrics to estimate the impact of the effects and found that each metric measures only one aspect of the impact and that conclusions based on one metric can be misleading.
This necessitates a more general approach to model the correlation between embedding deformation and effect strength.

Instead of studying the sensitivity of embeddings, other existing work focuses on their robustness.
Sensitivity is broader than robustness in the context of pre-trained embeddings.
The former measures the impact of any known factors on the embeddings, while the latter gauges the resilience of the embeddings to unwanted parameters for a specific downstream task. 
Abe{\ss}er et al.~\cite{abesser2023robust} explored the robustness of audio embeddings for polyphonic sound event tagging.
It was found in~\cite{wang2023transfer, bailey2021gender} that the downstream classification performance of pre-trained audio embeddings is not robust to dataset identity.
By extracting a dataset separation direction in the embedding space, this sensitivity could be potentially mitigated~\cite{wang2023transfer}. 
However, this quantification is based on the strong assumption that the bias subspace is low-dimensional and that dataset identity is linearly separable.
Indeed, only when this assumption holds can a (post-processing) projection reduce the unwanted sensitivity for a downstream task.

\begin{figure*}
    \centering
    \includegraphics[width=.765\linewidth]{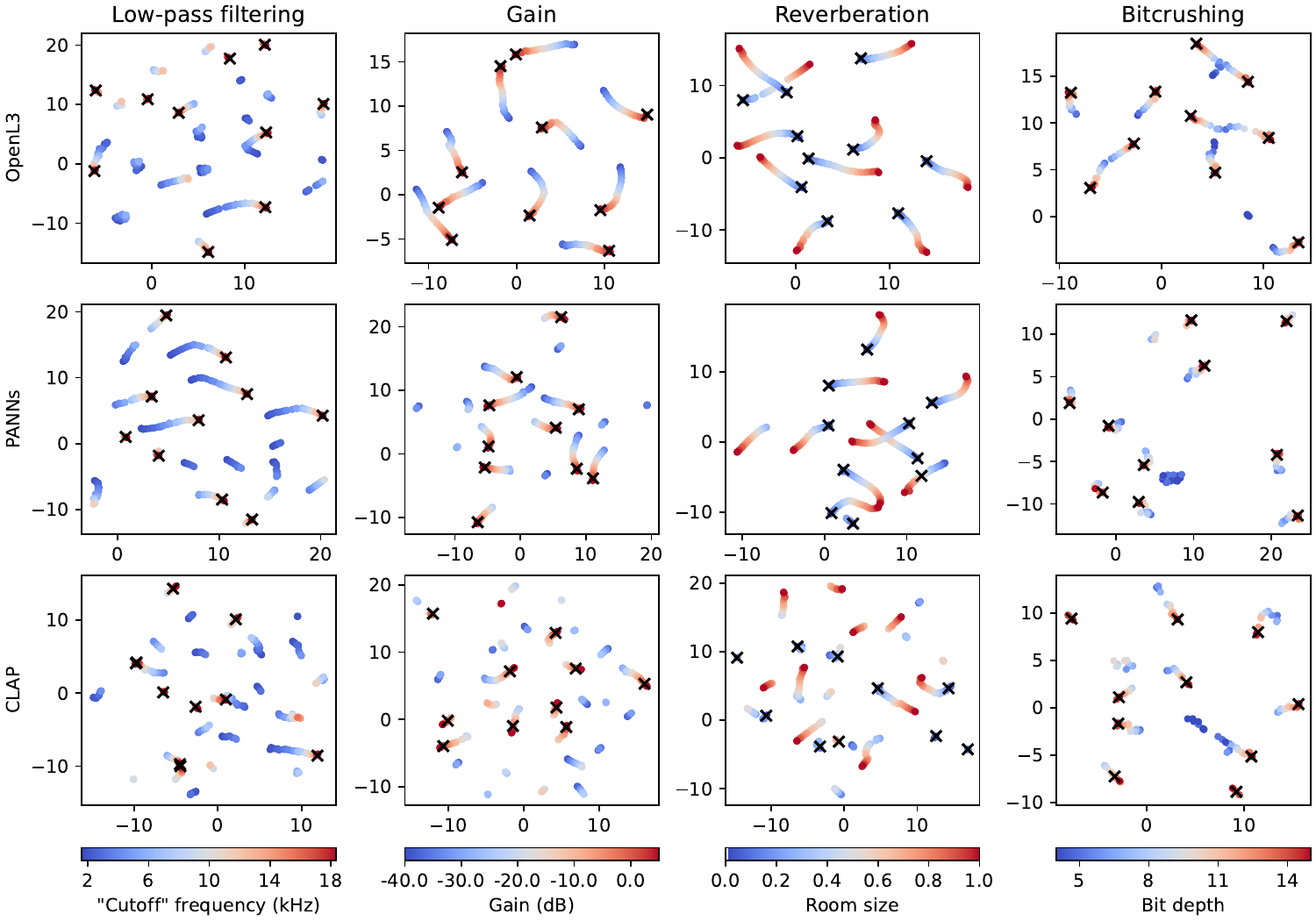}
    \caption{UMAP projected visualizations of the deformation trajectories of the embeddings after applying parameterized audio effects. Colormaps represent the effect strength. ``$\boldsymbol{\times}$" marks the embeddings of the original audio, while the colored points are those of the effected audio. }
    \label{fig:trajectory_plots}
\end{figure*}

To our knowledge, there is no existing research that systematically quantifies the sensitivity of pre-trained audio embeddings to known parameters and analyzes the properties of the resulting correlation.
We propose to introduce measurable impact by applying parameterized audio effects.
Effects are typical sources of sensitivity of audio embeddings as adding audio effects is a common data augmentation technique for training machine learning models~\cite{spijkervet2021contrastive, ramires2019data}.
Additionally, large audio datasets often consist of samples recorded under diverse conditions and may contain various audio effects. 
We investigate three main questions in this paper and open source the code\footnote{\url{https://github.com/vdng9338/audio-embedding-sensitivity}}.

\begin{itemize}
    \item How does an embedding represent a continuous deformation of an audio signal under common effects? (Section \ref{sec:impact})
    \item Is the response of an embedding to an effect consistent across audio examples and how to quantify the correlation between the response and the effect parameters? (Section \ref{sec:quantify})
    \item Can sensitivity to effects be neutralized by subspace projection methods if it is undesired for a downstream task? (Section \ref{sec:improve})
\end{itemize}

\section{Impact of audio effects}\label{sec:impact}

To understand how embeddings represent a continuous deformation, 
we apply common effects with a sweep of their parameters to the audio signal and visualize the response of the embeddings.
The structure of the deformation trajectories carved out by the parameter sweep is our core interest.

\subsection{Audio embeddings and audio effects}\label{sec:embedding_effect}
We consider three embedding models in this paper: OpenL3 (music/512), PANNs\cite{kong2020panns}, and CLAP~\cite{CLAP2022,CLAP2023}. 
OpenL3 is a pre-trained Look, Listen and Learn (L3) neural network trained on the task of audio-visual correspondence in a self-supervised manner.
CLAP\cite{CLAP2022,CLAP2023} and PANNs~\cite{kong2020panns} are two audio embedding models that achieve state-of-the-art
performance on various classification tasks and share the same architecture (CNN14) except that CLAP adds an extra linear projection layer.
Contrary to OpenL3 which produces frame-wise embeddings of dimension 512, CLAP and PANNs output a single embedding per file, with 1024 and 2048 dimensions, respectively. 

We include four simple and commonly-used audio effects: gain, low-pass filtering, reverberation, and bitcrushing which introduces distortion by reducing the resolution of the audio data. 
Each effect has a single key parameter of interest and represents qualities that many downstream tasks are generally invariant to.
We implement the effects using either the Pedalboard~\cite{sobot2023pedalboard}
or Scipy~\cite{virtanen2020scipy}
Python library; and employ the following parameter grid for each effect:
\begin{itemize}
    \item Gain: $[-40, +5]$ dB
    \item Low-pass filtering (Chebyshev type-II):
    cutoff frequencies $[1600, 18333]$ Hz
    \item Reverberation: room sizes $[0.01, 1.00]$ (normalized)
    \item Bitcrushing: bit depths $[4, 15]$
\end{itemize}
We apply the effects to audio examples in the IRMAS dataset, which gathers music excerpts from 11 instrument classes (see Section~\ref{sec:downstream}).


\subsection{Sensitivity of audio embeddings to audio effects}
\label{subsec:sensitivity}
We show how the effects impact the embeddings by visualizing the embedding response. We make use of the UMAP library~\cite{mcinnes2018umap-software} to obtain a 2-D representation of the audio embeddings. More precisely, fixing an audio embedding, audio effect and instrument, we perform the following:
Denote by $x^i$ the embedding of the $i$-th original audio frame ($i=1, 2, ..., N$), all belonging to the chosen instrument class, and by $x^i_p$ the embedding of the $i$-th audio frame to which the chosen effect at parameter $p$ was applied. 
We randomly choose 10 samples $i_1, i_2, ..., i_{10}$, we fit a UMAP projector on all the $(x^{i_k}_p)_{1 \leq k \leq 10,p}$ with a neighborhood size of 3, and we plot the UMAP projections of the $x^{i_k}$ and $(x^{i_k}_p)_p$ for $1 \leq k \leq 10$.
We use a neighborhood of size 3 because there is a single underlying degree of freedom in each parameter sweep; thus we should expect to observe 1-D manifolds in the embedding space that are connected by 3-nearest neighbors.

Fig.~\ref{fig:trajectory_plots} displays the embedding response visualizations of the three embedding models under the four effects presented in Section \ref{sec:embedding_effect}.
The audio examples are music excerpts of the cello instrument. 
Although these are a subset of examples, they are representative of the broader trends across embeddings, instruments and examples.
We summarize the following key observations per effect: 

\textbf{Low-pass filtering:} The response trajectories are generally not continuous, except for PANNs in some cases; the effected embeddings of each sample do not collapse to a small point cloud or trajectory at lower cutoff frequencies, but they do at higher ones.

\textbf{Gain:} The embedding trajectories are partially continuous for CLAP, and continuous in most cases for both PANNs and OpenL3.

\textbf{Reverberation:} Regardless of the foundation model, the trajectories are mostly continuous, except that the uneffected sample is sometimes separated from its effected versions. 

\textbf{Bitcrushing}: The behavior differs from model to model. For CLAP, the trajectories are rather discontinuous. For OpenL3, the trajectories are mostly continuous. For PANNs, the trajectories are short and the embeddings of the samples with a bit depth of 10 or more approximately almost collapse to a single point.

These observations suggest that the audio embeddings are sensitive to the audio effects, except for PANNs at high bit depths. 
More importantly, when the trajectories are continuous, they are also approximately linear.
This points towards the possibility that the audio embeddings linearize the effects at a sample-wise level.
However, the directions of the trajectories differ from sample to sample, suggesting that this linearization might not hold at a global level.
In terms of trajectory continuity, we notice that PANNs and OpenL3 yield more continuous trajectories than CLAP.
These interesting observations motivate us to quantitatively measure the impact of audio effects on pre-trained audio embeddings.

\section{Quantifying embedding sensitivity}\label{sec:quantify}

In this section, we investigate whether there is a single direction or a low-dimensional subspace in the embedding space that contains the deformation introduced by audio effects. 
As in \chang{S}ection \ref{subsec:sensitivity}, we fix an instrument and audio effect and assume that all the samples $i=1, ..., N$ belong to the fixed instrument class.

To find a potential deformation direction, we perform canonical correlation analysis (CCA) \cite{hotelling1992relations} between the embedding variables (that we will denote by $\xi_1, ..., \xi_d$) and the rank-transformed effect parameter (that we will denote by $y$), so as to find the direction in the embedding space that is most correlated with the effect strength. Mathematically, CCA between the random vector $\Xi = (\xi_1, ..., \xi_d)$ and the random variable $y$ consists in finding $u \in \mathbb{R}^d$ and $a \in \{-1, 1\}$ that maximize the correlation $\rho = \operatorname{corr}(u^\top \Xi, ay)$. For this computation, one can either consider all the data points $(x^i_p, y_p)$ for all $i$ and $p$, where $y_p$ denotes the rank of parameter $p$ -- we will refer to this as \emph{global CCA} --, or fix a sample $i$ and consider only $(x^i_p, y_p)$ for all $p$ -- we will refer to this as \emph{sample-wise CCA}.
To quantify the correlation between this direction and the effect parameter, in both the global and sample-wise cases, we plot the rank-transformed parameter $y$ against the scalar product of the effected embedding with the deformation direction, i.e. $\langle u, \Xi \rangle$, and then compute the squared Spearman correlation coefficient between these two variables that we will call $R^2$ coefficient.

\begin{figure}[t]
    \centering
    \includegraphics[width=\linewidth]{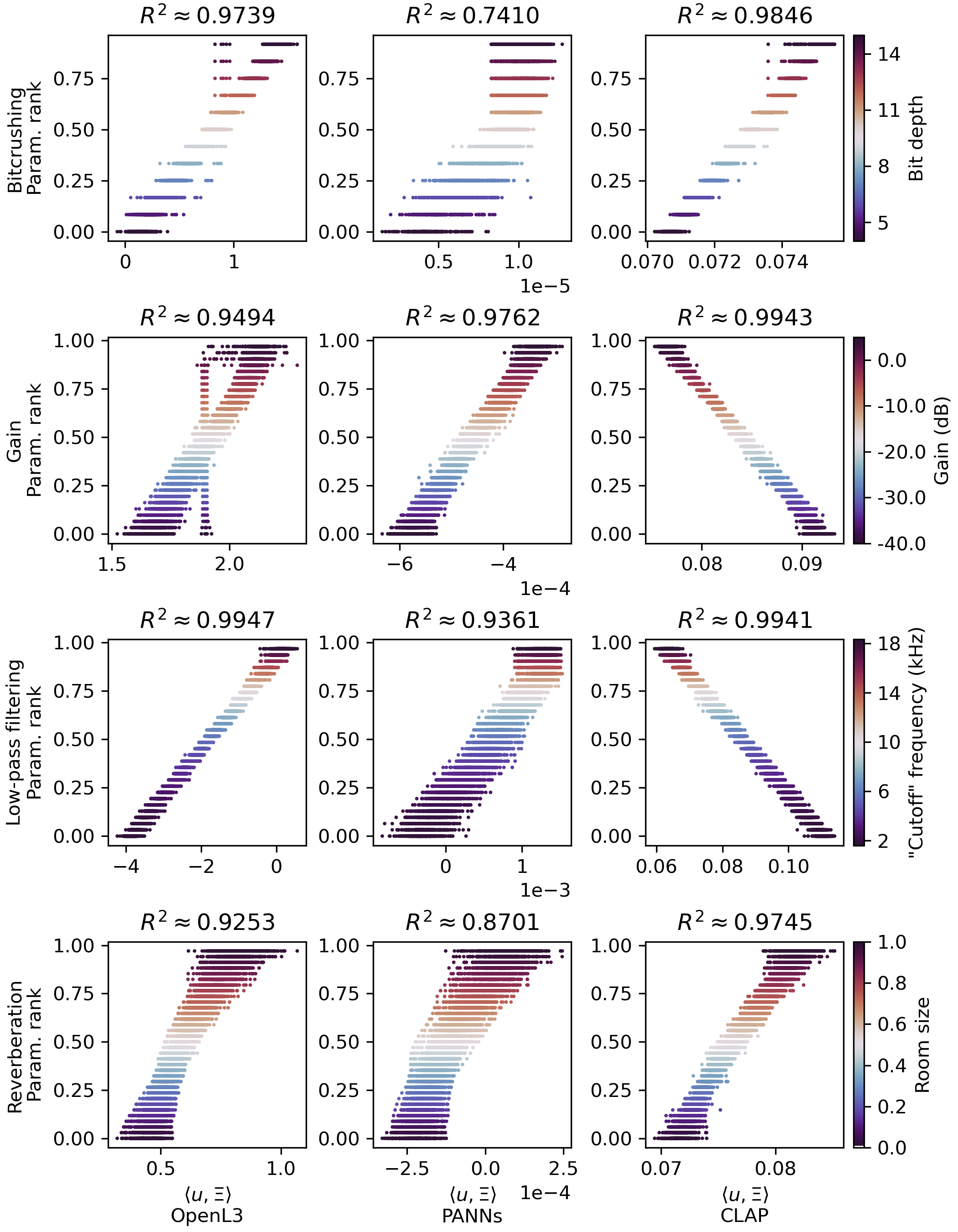}
    \caption{Correlation between the estimated deformation direction and effect strength for collections of audio samples. Cello samples.
    }
    \label{fig:correlation}
\end{figure}

Fig.~\ref{fig:correlation} shows global CCA correlation plots and corresponding $R^2$ coefficients for all combinations of embeddings and audio effects for the cello instrument. Table~\ref{tab:global_cca_corr_coefs} summarizes the different correlation coefficients for each combination of audio effect, audio embedding and instrument. For OpenL3 and CLAP, the correlation coefficients are almost always above 0.95 (with one exception for OpenL3 and reverberation, where the correlation coefficients are still above 0.9), and for PANNs, the correlation coefficients are almost always near or above 0.9, meaning that for all embeddings and most audio effects studied, there is a direction in the embedding space that correlates highly with the audio effect strength, though this does not strongly hold for PANNs. Note that bitcrushing with PANNs is an exception here; when plotting the distance matrices of the PANNs embeddings of bit-crushed samples, we found a clustering of embeddings at bit depths higher than 10 approximately.

\begin{table}[hbt!]
	\centering
         \normalsize
          \scalebox{0.87}{
	\begin{tabular}{c|c|c|c|c}
		& Low-pass filt. & Reverb. & Bitcrushing & Gain \\
	   \hline
	   OpenL3 & $99.42 \pm 0.13$ & $93.83 \pm 1.11$ & $98.30 \pm 0.47$ & $96.46 \pm 0.95$ \\
	   PANNs & $92.64 \pm 1.18$ & $89.06 \pm 2.19$ & $69.64 \pm 7.87$ & $97.90 \pm 0.49$ \\
	   CLAP & $99.27 \pm 0.12$ & $98.09 \pm 0.36$ & $98.68 \pm 0.35$ & $99.63 \pm 0.12$ \\
	\end{tabular}}
	\caption{Instrument-wise global CCA correlation coefficient statistics (mean $\pm$ standard deviation) for each embedding and each audio effect. All numbers are multiplied by $10^2$.}
	\label{tab:global_cca_corr_coefs}
\end{table}

However, having a high global $R^2$ coefficient does not necessarily mean that the deformation induced by the audio effect is one-dimensional. CCA can indeed find a correlation coefficient of 1 with any trajectory or set of trajectories that are monotonous along some direction, no matter the variations of the trajectories in other directions. To check whether the deformation induced by the audio effect is globally low-dimensional, we compute all the sample-wise CCA directions and perform singular value decomposition (SVD) on them, then compare the singular values of the sample-wise CCA directions with those of the (centered) original embeddings. If this comparison exhibits a high dimensionality of the sample-wise CCA directions, it would confirm that the deformation induced by the audio effect is high-dimensional though the converse may not be true. 

\begin{figure}[t]
    \centering
    \includegraphics[width=.95\linewidth]{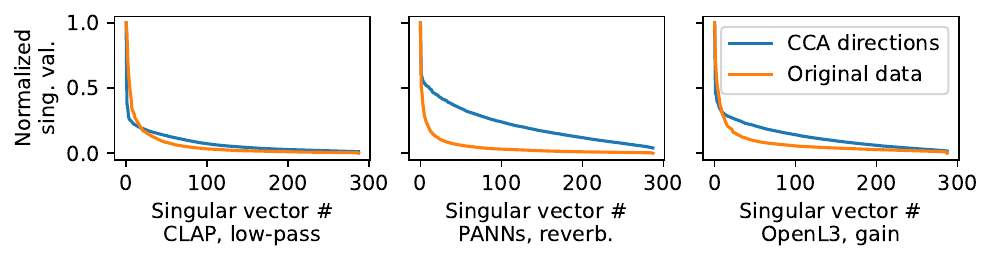}
    \caption{Comparison of singular values of SVD of sample-wise CCA \chang{directions} and PCA explained variances of original data for cello and three combinations of embedding and audio effect.}
    \label{fig:svd_samplewise}
\end{figure}

Fig.~\ref{fig:svd_samplewise} shows example comparisons of the singular values of the sample-wise CCA directions and the centered original data for some combinations of embedding models, audio effects and instruments; these examples are representative of most such combinations. The plotted singular values for each curve are divided by the largest singular value. In all cases, the comparisons exhibit a high dimensionality of the sample-wise CCA directions even though this dimensionality appears to be slightly lower in some cases with OpenL3, like with the low-pass filtering effect. 
This shows that the deformations induced by the four audio effects considered are not linear or low-dimensional.

To check whether the deformation induced by the audio effect can be locally linear, we check whether the $R^2$ coefficients of the sample-wise CCAs are high. It turns out that for \emph{all} combinations of audio effect and embedding and \emph{all} samples, the $R^2$ coefficient is equal to 1.
This demonstrates that for all audio samples, there exists a direction along which the embeddings move monotonically as the audio effect strength increases (possibly but not necessarily linearly). 


\section{Reducing embedding sensitivity}\label{sec:improve}

\subsection{Methods}
With the deformation direction identified, we explore the possibility of reducing the sensitivity of the embeddings to audio effects.
We propose to project out this direction or an estimated deformation subspace in a broader sense in the embedding space.
We study four methods apart from global CCA to estimate the deformation direction or subspace:

\textbf{Sample-wise CCA SVD:} 
For each sample $i$ of the same instrument class, performing CCA between $(x^i_p)_p$ and $(y_p)_p$ yields correlation direction $c_i$. Then, we perform SVD of the $c_i$, yielding right singular vectors $u_1, u_2, ..., u_K$ associated to singular values $s_1 \geq s_2 \geq ... \geq s_K \geq 0$. We fix a threshold $t \in [0, 1]$ and project out all the $u_k$ such that $s_k \geq ts_1$. $t=0.3$, $t=0.4$ and $t=0.5$ are used.

\textbf{Principal component analysis:} We perform PCA of the displacements $(x^i_p-x^i)_{p\ \text{not neutral}}$ for all $i$ and keep the principal component with the largest explained variance (absolute or relative sense) as the deformation direction: letting $u_1, u_2, ..., u_K$ be the principal components of the displacements, $\sigma_1^2 \geq ... \geq \sigma_d^2 \geq 0$ the corresponding explained variances, and $\tau_1^2 \geq ... \geq \tau_K^2 \geq 0$ the explained variances of the PCA of the uneffected embeddings, we either project out $u_1$ (absolute sense), or we project out the principal component $u_i$ that maximizes the ratio $\sigma_i^2/\tau_i^2$ (relative sense). 

\textbf{Average displacement:} 
We consider the normalization to unit length of $\frac1N \frac1P \sum_{p\ \text{not neutral}} (x^i_p-x^i)$ as the \chang{deformation direction}.

\textbf{Linear discriminant analysis:} We perform linear discriminant analysis (LDA) between two classes of points: the first class of points contains all the $x^i$ for all $i$, the second class of points contains all the $x^i_p$ for all $i$ and $p$ such that $p$ is non-neutral. LDA yields a discriminant direction $w$ and a constant $c$ such that under some assumptions, a point $x$ is more likely to belong to the second class if and only if $w^\top x > c$; we project out $w/\lVert w \rVert$.

\subsection{Evaluation}\label{sec:downstream}

\noindent \textbf{Downstream task: music instrument classification. }
For each sensitivity mitigation method and each instrument class, we train a logistic regressor on the task of recognizing the instrument on the desensitized embeddings of some training datasets (effected or uneffected) and test it on those of some test datasets; we use ROC AUC for evaluation. These ROC AUCs are compared to those of the classifier trained and tested with the embeddings whose sensitivity has not been reduced. 

\vspace{.5ex}
\noindent \textbf{Dataset} We use the 
IRMAS dataset~\cite{bosch2012comparison} for experiments.
The dataset contains 6705 music excerpts of 11 instrument classes.
Each excerpt is 3 seconds in length.

For each parameter of each audio effect, we perform an experiment where we train the logistic classifier on the uneffected dataset and test on the effected dataset at this particular parameter, and another experiment where we swap the train and test sets.


\subsection{Results}


\begin{figure}[t]
    \centering
    \includegraphics[width=\linewidth]{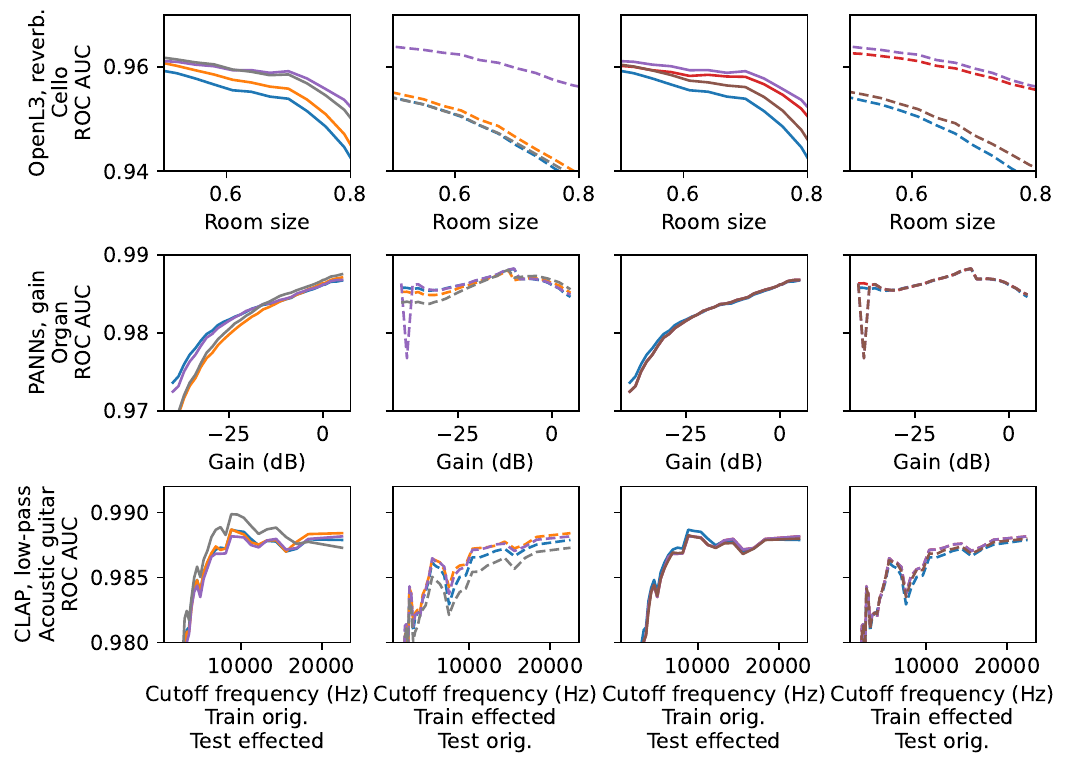} \\
    \includegraphics[width=.4\linewidth]{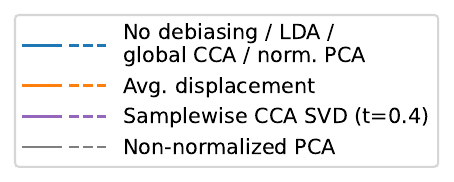}
    \includegraphics[width=.4\linewidth]{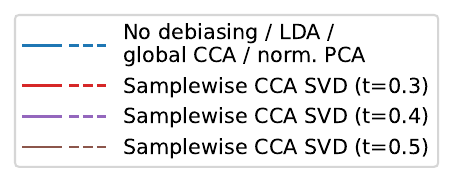}
    \caption{Classification performance \chang{comparison (debiasing is equivalent to desensitizing here) in terms of ROC AUC}. 
    }
    \label{fig:classification_results}
\end{figure}

Fig.~\ref{fig:classification_results} shows classification ROC AUCs for three combinations of audio embeddings, audio effects and instruments, which are representative of most combinations. 
Without sensitivity reduction, the classification performance is usually sensitive to the audio effect strength, with a higher effect strength inducing a sharper drop in classification performance. In most cases, with sensitivity reduction, the classification performance remains sensitive to the audio effect strength, which confirms that the deformation induced by the audio effects is not one-dimensional (in the case of one-dimensional projections) or low-dimensional (in the case of the sample-wise CCA SVD method). However, there are minor variations in classification performance compared to that without sensitivity reduction, in both directions (higher and lower performance). 

We can make the following general observations: (1) Global CCA projection and LDA projection have virtually no impact on the classification performance in most cases (overlapped curves without sensitivity reduction are not plotted for clarity).
(2) In many cases, average displacement projection improves classification performance by around 0.003 to 0.01 AUC (Fig. \ref{fig:classification_results} top), but it sometimes decreases the performance by the same order of magnitude (Fig. \ref{fig:classification_results} middle), and in many other cases, the impact of average displacement projection on the classification performance is neutral (Fig. \ref{fig:classification_results} bottom, to some extent). The same can be said for the non-normalized PCA projection variant and sample-wise CCA SVD projection, except that cases where the performance decreases are more common; a threshold of $t=0.4$ seems to more often perform better than $t=0.3$ and $t=0.5$ in Fig. \ref{fig:classification_results}, but this depends on the cases, with $t=0.3$ performing better in many other cases.
(3) With PANNs and CLAP (and all four audio effects), and with reverberation and OpenL3, the normalized PCA projection variant has a neutral effect on the classification performance. With OpenL3 and the three other audio effects (not plotted here), we observe a behavior similar to (2).

\section{Conclusion}

We propose a framework to quantify the sensitivity of pre-trained audio embeddings to common effects. By applying parameterized audio effects, we analyze the correlation between the embedding response and the effect strength, and derive an estimated deformation direction. 
Our findings indicate that the deformation subspace is generally high-dimensional, suggesting that embeddings do not linearize audio effects in the embedding space. Consequently, a linear post-processing approach, i.e. projecting out the deformation direction or subspace, may hardly improve the robustness of pre-trained audio embeddings to effects for downstream tasks. 
The proposed pipeline could be potentially generalized to analyze the sensitivity of any foundation models to any known parameters, beyond audio effects.

\bibliographystyle{IEEEtran}
\bibliography{refs}

\end{document}